\documentclass[sigconf]{acmart}
\copyrightyear{2026}
\acmYear{2026}
\setcopyright{cc}
\setcctype{by}
\acmConference[MM '26] {Proceedings of the 34th ACM International Conference on Multimedia}{November 10--14, 2026}{Rio de Janeiro, Brazil.}
\acmBooktitle{Proceedings of the 34th ACM International Conference on Multimedia (MM '26), November 10--14, 2026, Rio de Janeiro, Brazil}
\acmISBN{979-8-4007-2213-4/2026/11}
\acmDOI{10.1145/3767308.3835111}
\AtBeginDocument{%
  }
\usepackage{multirow} 
\usepackage{siunitx}
\begin{document}

\title{Restoring Semantic Fidelity for Training-Free High-Resolution Image Generation in Diffusion Transformers}

\author{Junjie Li}
\authornote{Both authors contributed equally to this research.}
\orcid{0000-0002-2906-8514}
\affiliation{%
  \institution{Beijing University of Posts and Telecommunications}
  \city{Haidian}
  \state{Beijing}
  \country{China}
}
\email{2010919504@bupt.cn}

\author{Zengbin Wang}
\orcid{0000-0002-9319-905X}
\authornotemark[1]
\affiliation{%
  \institution{Beijing University of Posts and Telecommunications}
  \city{Haidian}
  \state{Beijing}
  \country{China}}
\email{wzb1@bupt.edu.cn}

\author{Gaoxiang Li}
\orcid{0000-0003-4262-7699}
\affiliation{%
  \institution{Beijing Normal University}
  \city{Haidian}
  \state{Beijing}
  \country{China}
}
\email{202021210031@mail.bnu.edu.cn}
\author{Muyi Sun}
\orcid{0000-0001-9506-7643}
\affiliation{%
 \institution{Beijing University of Posts and Telecommunications}
  \city{Haidian}
  \state{Beijing}
  \country{China}
\email{muyi.sun@bupt.edu.cn}
}
\author{Yixin Zhang}
\orcid{0000-0002-4513-1106}
\authornote{Project Leader.}
\affiliation{%
  \institution{Institute of Artificial Intelligence, Hefei Comprehensive National Science Center}
  \city{Hefei}
  \state{Anhui}
  \country{China}}
\email{zhyx12@ustc.edu.cn}

\author{Man Zhang}
\orcid{0000-0003-3043-2122}
\correspondingauthor
\authornote{Corresponding Author.}
\affiliation{%
  \institution{Beijing University of Posts and Telecommunications}
  \city{Haidian}
  \state{Beijing}
  \country{China}}
\affiliation{%
     \institution{Qinghai Institute of Technology}
  \city{Xining}
  \state{Qinghai}
  \country{China}}
\email{zhangman@bupt.edu.cn}
\renewcommand{\shortauthors}{Junjie Li et al.}

\begin{abstract}
Diffusion Transformer (DiT)-based text-to-image models have recently achieved remarkable success in image synthesis. However, when directly applied beyond their training resolution, these models often suffer from severe semantic degradation, where meaningful semantic content disappears. Although existing training-free high-resolution generation methods can mitigate this issue, the inherent semantic degradation at high resolutions remains largely unresolved, leading to degraded fine-grained details. Furthermore, widely used structural guidance methods lack modeling of the structure and details in the intermediate trajectory, which degrades the quality of the generated results. In this paper, we propose a training-free framework for high-resolution image generation to address these challenges. First, we propose a text–visual scale-decoupled attention (TVSDA) module, which restores the model’s ability to generate semantically meaningful content at high resolutions by preserving the global receptive field of text tokens while aligning the receptive field of image tokens with the training resolution. Second, we propose a structure–detail guidance (SDG) strategy that explicitly models structural information and fine details through scale-wise and temporal-wise guidance, overcoming the limitations of conventional low-resolution guidance. Extensive experiments demonstrate that our method significantly improves both structural coherence and detail fidelity over existing training-free DiT-based high-resolution generation approaches. The code is available at \url{https://github.com/ljjcoder/Restoring-Semantic-Fidelity}.

\end{abstract}




\maketitle

\section{Introduction}
\label{sec:intro}

\begin{figure}
    \centering
    \includegraphics[width=1.0\linewidth]{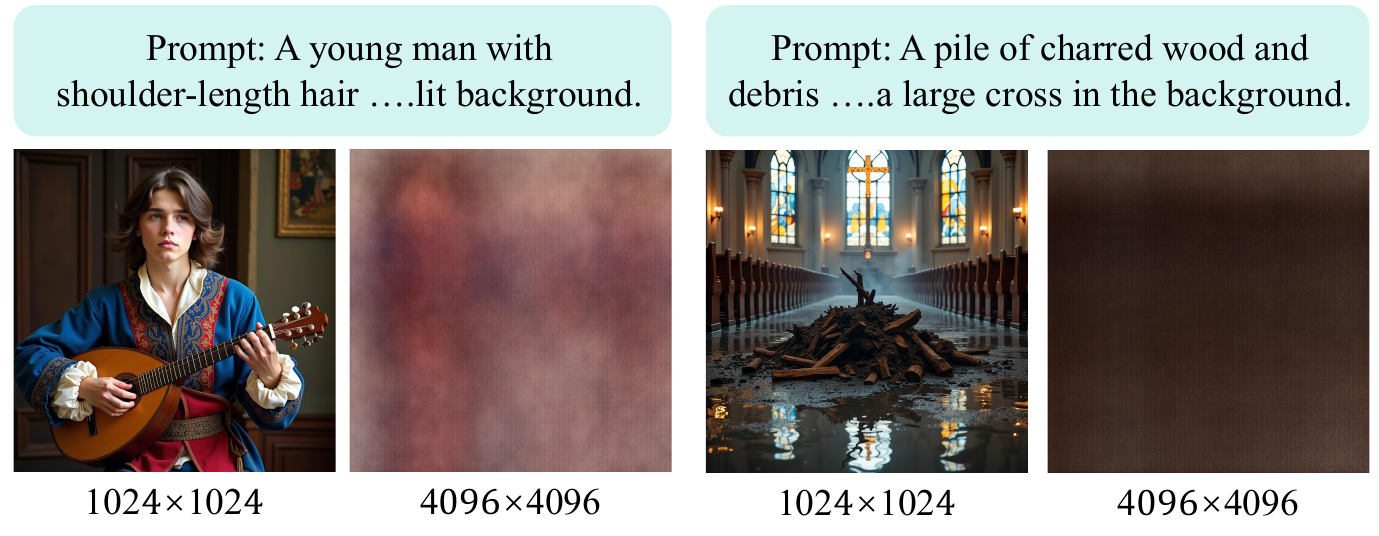}
    \vspace{-0.25cm}
    \caption{Illustrations of semantic degradation. When performing high-resolution inference with FLUX~\cite{flux2024} (trained at 1024$\times$1024 and tested at 4096$\times$4096), semantic structures completely disappear. \textbf{Zoom in} for better visualization.}
    
    \label{figure:hg_issue}
    \vspace{-5pt}
\end{figure}

In recent years, text-to-image generation models~\cite{flux2024,esser2024scaling} based on Diffusion Transformers (DiT)~\cite{peebles2023scalable} have substantially advanced image synthesis with remarkable visual quality. However, when the target resolution exceeds the training resolution, these models often fail dramatically~\cite{du2024max,bu2025hiflow}: meaningful semantic content disappears, and the generated images may collapse into semantically unrecognizable patterns, as shown in Figure~\ref{figure:hg_issue}. This phenomenon, which we refer to as \emph{semantic degradation}, has become a major obstacle for high-resolution generation scenarios.

A straightforward solution is to further fine-tune pre-trained models on high-resolution images~\cite{guo2024make,hoogeboom2023simple,liu2024linfusion,ren2024ultrapixel,zhang2025diffusion,yu2025ultra}. However, this strategy is often impractical due to the high computational cost and the limited availability of large-scale high-resolution training data. As a result, training-free high-resolution generation has attracted increasing attention~\cite{he2023scalecrafter,du2024demofusion,kim2025diffusehigh,bar2023multidiffusion,lee2023syncdiffusion,du2024max,bu2025hiflow}. Early approaches were mainly developed for U-Net architectures~\cite{he2023scalecrafter,zhang2023hidiffusion,huang2024fouriscale}, and are not readily applicable to DiT-based models. More recent studies~\cite{du2024max,bu2025hiflow} have begun to explore training-free high-resolution inference for DiTs, significantly improving generation quality.

\begin{figure}
    \centering
    \includegraphics[width=1.0\linewidth]{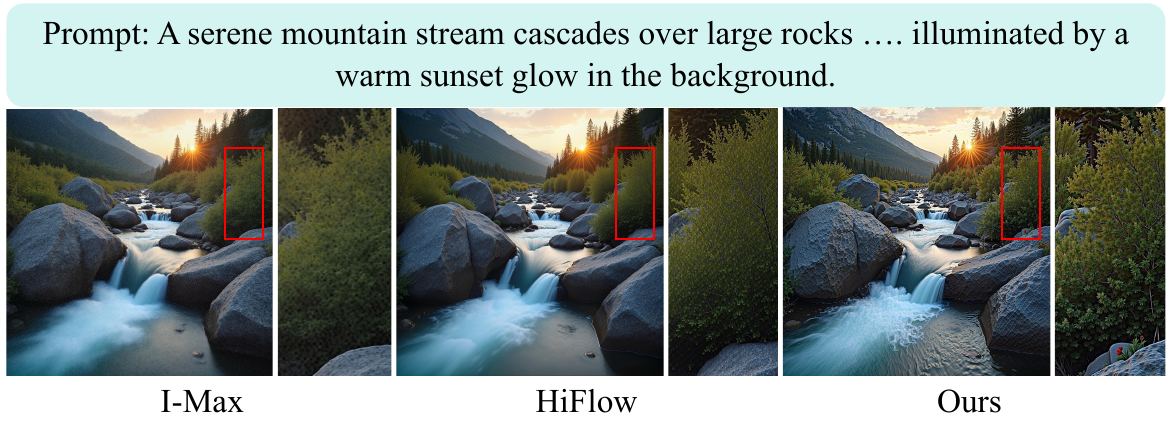}
    \vspace{-0.25cm}
    \caption{Visual comparison at 4096$\times$4096 resolution. DiT-based methods I-Max~\cite{du2024max} and HiFlow~\cite{bu2025hiflow} suffer from blurry backgrounds and a lack of high-frequency details, whereas our method restores fine-grained structures and produces sharper results. \textbf{Zoom in} for better visualization.}
    \label{figure:top_vis_issue}
    \vspace{-5pt}
\end{figure}

Despite recent progress, existing methods~\cite{du2024max,bu2025hiflow} primarily rely on low-resolution structural guidance, introducing structural priors during high-resolution inference rather than directly addressing the model's intrinsic semantic degradation. If the model lacks the ability to generate semantically consistent content, such external guidance alone is insufficient to produce fine-grained details, as shown in Figure~\ref{figure:top_vis_issue}. To address this issue, we propose a text-visual scale-decoupled attention (TVSDA) module to alleviate the model's intrinsic semantic degradation at high resolutions.
The core idea is to align the attention scale at high resolutions with that used during training. Concretely, attention for text tokens, which lack explicit positional information, is computed in a global manner, whereas attention for visual tokens with explicit positional information is confined within the training resolution. 

However, restoring semantic fidelity alone is not sufficient. Although TVSDA enables the model to generate semantically meaningful content, the resulting images often suffer from disorganized global layouts, manifested as repetitive local structures (see Figure~\ref{figure:swin_initial_issue}). A natural next step is therefore to introduce a coarse structural prior, as in existing methods~\cite{du2024demofusion,kim2025diffusehigh,qiu2024freescale,bu2025hiflow}, to improve the global layout. Nevertheless, it still leaves two important residual failures (see Figure~\ref{figure:pa_intial_issue}): \textbf{(1)} local semantic repetition, and \textbf{(2)} blurred fine details. These observations suggest that the remaining errors are no longer caused simply by a missing initial structure, but by a mismatch in how the reverse flow trajectory evolves at high resolution.

This leads to the important question of our work: \emph{how should the high-resolution flow trajectory be corrected once semantic fidelity has been restored and a coarse structure has been initialized?} To answer this, we perform an empirical analysis of the single-step predicted clean images across both different resolutions and different denoising timesteps  (see Figure~\ref{figure:cross_scale_time_motivation}). We observe that lower-resolution predictions tend to provide more stable global semantic layouts, whereas higher-resolution predictions contain richer high-frequency details. Moreover, as denoising proceeds, later predictions become progressively sharper and reveal increasingly clear fine details. These observations suggest that structural coherence and fine-detail restoration should be guided in two ways: across scales and across timesteps.

Based on this analysis, we propose a \emph{structure--detail guidance} (SDG) strategy to refine the velocity field throughout inference. SDG consists of two key components. The first is \emph{Cross-Scale Guidance} (CSG), which encourages the low-frequency components of the current high-resolution prediction to align with the structurally stable low-resolution prediction, while preserving the high-frequency components from the high-resolution branch. The second is \emph{Cross-Temporal Guidance} (CTG), which leverages the progressive emergence of high-frequency details along the denoising trajectory to further sharpen fine structures. Ultimately, as shown in Figure~\ref{figure:top_vis_issue}, with the integration of TVSDA and SDG, our method not only preserves coherent semantic layouts but also significantly enhances fine-grained details compared to previous methods.

The contributions of this paper are summarized as follows.
\begin{itemize}
\item We propose a text--visual scale-decoupled attention (TVSDA) module to restore semantic fidelity by preserving global text interactions while constraining visual attention to training local regions.

\item We show that conventional low-resolution structural guidance is a natural but insufficient remedy: although it improves coarse layout, it still leaves local semantic repetition and blurred details. To address these residual failures, we analyze predicted clean images across scales and timesteps, and derive a structure--detail guidance strategy consisting of Cross-Scale Guidance (CSG) and Cross-Temporal Guidance (CTG) to refine the high-resolution flow trajectory.

\item Extensive experiments demonstrate that our method consistently outperforms prior training-free DiT-based approaches in both generation quality and inference efficiency on high-resolution image generation tasks.
\end{itemize}
\section{Related Work}
\label{sec:related_work}
\subsection{Text-to-Image Generation}
Diffusion models~\cite{blattmann2023align,ho2020denoising,wang2024exploiting,song2020denoising}  have emerged as the prevailing framework for text-to-image generation~\cite{rombach2022high,podell2023sdxl}, owing to their strong capability in modeling complex high-dimensional distributions. Early works~\cite{ho2020denoising,ho2022classifier,dhariwal2021diffusion} performed denoising at the pixel level, resulting in significantly high computational costs during both training and inference. The Latent Diffusion Model (LDM)~\cite{rombach2022high} was subsequently introduced to mitigate these costs, enabling more efficient training and inference while enhancing the model’s ability to generate high-resolution images.

Despite these advances, the training resolution of current diffusion models remains well below that required for many real-world applications. When the target resolution exceeds the training resolution, generation quality deteriorates substantially. Fine-tuning with high-resolution images~\cite{ren2024ultrapixel,zhang2025diffusion,yu2025ultra} can address this issue, but it is computationally expensive and limited by the availability of high-resolution data.

\subsection{Training-Free High-Resolution Image Generation}
To address these issues, several works~\cite{he2023scalecrafter,du2024demofusion,lee2023syncdiffusion,du2024max,bu2025hiflow} explore strategies to leverage pre-trained U-Net-based models (e.g., SDXL~\cite{podell2023sdxl}) for high-resolution image generation without additional training.
These methods can be broadly categorized into three types: (1) patch-wise methods~\cite{bar2023multidiffusion,lee2023syncdiffusion}, which divide the input into overlapping patches, perform standard denoising on each patch, and then fuse the results; (2) model component optimization methods~\cite{he2023scalecrafter,zhang2023hidiffusion,huang2024fouriscale,jin2023training}, which adapt selected components of the pre-trained model—such as self-attention layers~\cite{jin2023training} or convolutional layers with dilated convolutions~\cite{he2023scalecrafter,zhang2023hidiffusion,huang2024fouriscale}—for high-resolution generation; and (3) structural guidance methods~\cite{du2024demofusion,kim2025diffusehigh,qiu2024freescale},  which leverage low-resolution images to provide semantic guidance for high-resolution generation. 

Recently, transformer-based diffusion models~\cite{flux2024,gao2025lumina,esser2024scaling} have significantly outperformed U-Net–based architectures. Consequently, several training-free high-resolution generation methods~\cite{du2024max,bu2025hiflow,koh2025scalediff,zhou2026lookflow,yu2026realuhr,ma2025resdit}  built on the transformer architecture have been proposed, which improve generation quality. However, these methods still tend to produce blurry results.


In this work, we focus on high-resolution image generation using the transformer architecture. Unlike previous approaches, our method enhances generation quality by combining two strategies: restoring the model’s generative capabilities and refining the reverse denoising trajectory with structural and detail guidance.


\section{Background}
\label{sec:Background}
Before introducing our method, we briefly review the pretrained model, including its training and inference strategies as well as its architecture.

\subsection{Training-Inference Process}
Current advanced generative models~\cite{flux2024,esser2024scaling} generally employ rectified flow~\cite{liu2022flow}  to model the trajectory between noisy and clean images. Rectified Flow attempts to simplify the training objective of DDPM~\cite{ho2020denoising} by establishing the transport trajectory between the two distributions through linear interpolation. Given clean image samples $x_0 \sim \pi_0$ and Gaussian noise $x_1 \sim \pi_1$, Rectified Flow (RF) constructs a linear interpolation path from $x_1$ to $x_0$. The intermediate state at time $t$ is given by: 
\begin{equation}\label{eq:flowmatch}
    x_t = tx_1 + (1-t)x_0,
\end{equation}
where $t \in \left [0, 1  \right ] $ represents continuous time. The instantaneous velocity along this path is given by the time derivative:
\begin{equation}\label{eq:vexpress}
v(x_t, t) = \frac{d x_t}{d t} = x_1 - x_0 .
\end{equation}

RF aims to learn a time-dependent velocity field $v_\theta(x, t)$ that matches the target velocity along the interpolation path.
The training objective is defined as:
\begin{equation}
\label{eq:fm_objective}
\mathcal{L}(\theta)
=
\mathbb{E}
\left[
\left\|
v_\theta(x_t, t) - (x_1 - x_0)
\right\|_2^2
\right],
\end{equation}
where $x_0 \sim \pi_0$, $x_1 \sim \pi_1$, and $t \sim \mathcal{U}(0,1)$.

Once training is complete, clean images can be generated step by step as follows:
\begin{equation}\label{eq:v-theta_3}
    x_{t_{i-1}} = x_{t_{i}} + v_\theta(x_{t_{i}}, t_i)({t_{i-1}}-{t_{i}}),
\end{equation}
in which the movement of $t_i$ from 1 to 0 indicates the trajectory of $x_{t_{i}}$ from gaussian noise $x_1$ to clean sample $x_0$.
 In the following sections, $v_t$ is used to denote $v_\theta(x_{t}, t)$ for simplicity.

\subsection{Architecture}
Unlike previous U-Net-based models~\cite{rombach2022high,podell2023sdxl}, modern generative models~\cite{flux2024,esser2024scaling} adopt the more expressive Multimodal Diffusion Transformer (MMDiT)~\cite{peebles2023scalable}.
The MMDiT architecture consists of modules such as FFN, adaLN-Zero blocks, and joint self-attention. 
Among these, only the self-attention module is resolution-dependent; the other modules operate on a per-token basis and are therefore resolution-agnostic. Therefore, we next describe in detail the joint self-attention mechanism for text and visual tokens.

Let $Q_{\mathrm{v}}$ ($Q_{\mathrm{t}}$), $K_{\mathrm{v}}$ ($K_{\mathrm{t}}$), and $V_{\mathrm{v}}$ ($V_{\mathrm{t}}$) denote the query, key, and value matrices of image and text tokens, respectively. 
In standard self-attention, image and text tokens share the same keys and values:
\begin{equation}
\label{eq:joint_kv}
K = \left[  K_{\mathrm{t}}; K_{\mathrm{v}} \right], 
\qquad
V = \left[  V_{\mathrm{t}}; V_{\mathrm{v}} \right],
\end{equation}
where $[\,\cdot\,;\cdot\,]$ denotes concatenation along the token dimension.
This allows any query token—whether from text ($q^i \in Q_{\mathrm{t}}$) or image ($q^i \in Q_{\mathrm{v}}$)—to attend jointly to all tokens.

Formally, for a query token $q^i$, the attention output is computed as
\begin{equation}
\label{eq:standard_attention}
\mathrm{Attn}\!\left(q^i\right)
=
\sum_{j}
\mathrm{softmax}\!\left(
\frac{q^i K_{j}^{\top}}{\sqrt{d}}
\right)
V_{j},
\end{equation}
where $d$ denotes the channel dimensionality of the query embeddings.

\section{Method}
\label{sec:Method}
As illustrated in Figure~\ref{figure:hg_issue},  when the target resolution exceeds the training resolution, the pretrained model often produces images that lack meaningful semantic content.

\begin{figure*}
    \centering
    \includegraphics[width=0.98\linewidth]{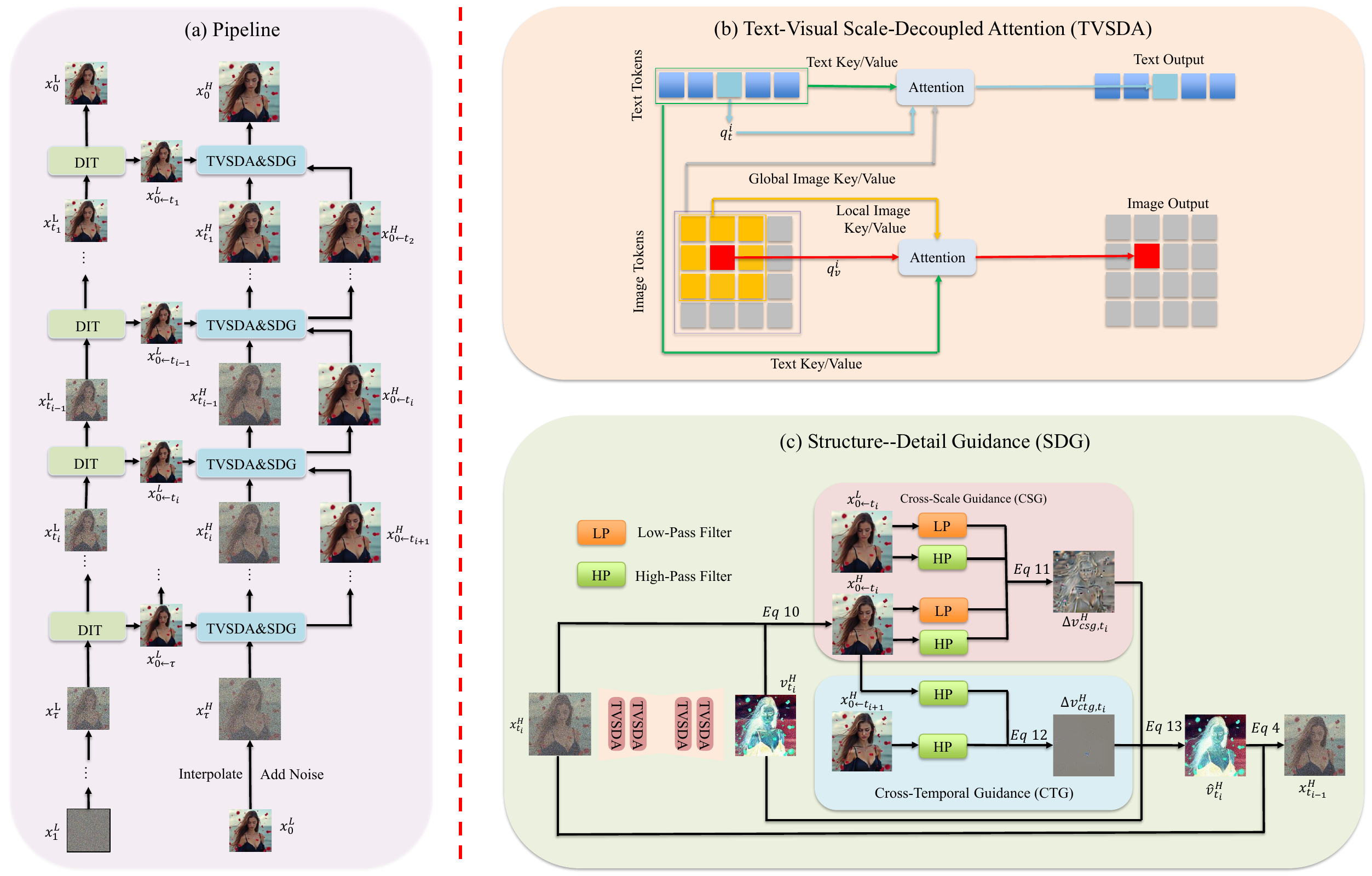}
    \vspace{-0.25cm}
    \caption{Pipeline of our method. During the DiT denoising process, Text-Visual Scale-Decoupled Attention (TVSDA) replaces the original attention to restore semantic generation capability, while Structure–Detail Guidance (SDG) preserves global structure and restores high-frequency details.}
    \label{figure:Framework}
    \vspace{-5pt}
\end{figure*}

To address this limitation, we introduce a text–visual scale-decoupled attention module and a structure–detail guidance strategy. The former reformulates the self-attention mechanism to restore the model’s ability to generate meaningful semantic content, whereas the latter leverages cross-scale and cross-temporal guidance to maintain structural coherence while enhancing high-frequency details.
%
 %
 An overview of the framework is illustrated in Figure~\ref{figure:Framework}.
 
\begin{figure}
    \centering
    \includegraphics[width=1.0\linewidth]{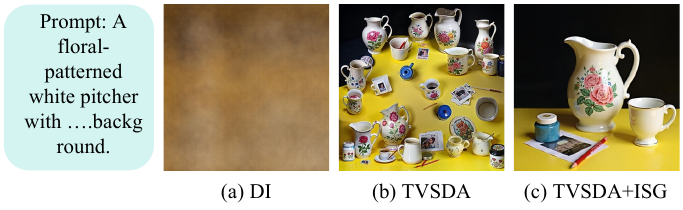}
    \vspace{-0.25cm}
    \caption{Comparison of high-resolution generation results (DI: direct inference). \textbf{Zoom in} for better visualization.
 } 
    \label{figure:swin_initial_issue}
    \vspace{-5pt}
\end{figure}



\subsection{Text-Visual Scale-Decoupled Attention}
\label{sec:tvsda}


In the joint self-attention layer, as the resolution increases, the receptive field of each query token grows quadratically~\cite{jin2023training}, making attention overly uniform at high resolutions, causing it to attend to many irrelevant regions, and ultimately leading to semantic degradation.


To mitigate this issue, we aim to align attention computation with the training resolution during high-resolution generation. For a visual query token $q_{\mathrm{v}}^i$, which is associated with an explicit spatial position, it is natural to constrain its receptive field to the local region centered on itself, with a scale consistent with the training resolution. In contrast, a text query token $q_{\mathrm{t}}^i$, lacking explicit positional priors,  is not constrained in the same way. Motivated by this, we propose a text-visual scale-decoupled attention mechanism (TVSDA), which decouples attention scales for text and visual tokens.



Specifically, attention is computed as follows:

\begin{enumerate}
    \item \textbf{Text queries} $q_{\mathrm{t}}^i$ follow standard global attention (Equation~\ref{eq:standard_attention}), 
    using the full set of keys and values.
    
    \item \textbf{Visual queries} $q_{\mathrm{v}}^i$ use the full text keys $K_{\mathrm{t}}$ and values $V_{\mathrm{t}}$, 
    while the visual keys and values are restricted to a local neighborhood centered at $q_{\mathrm{v}}^i$:
    \begin{equation}
        \tilde{K}_{\mathrm{v}}^i = K_{\mathrm{v}}[\mathcal{N}(i,s)], \quad 
        \tilde{V}_{\mathrm{v}}^i = V_{\mathrm{v}}[\mathcal{N}(i,s)],
    \end{equation}
    where $\mathcal{N}(i,s)$ denotes the set of visual tokens within a neighborhood of training scale $s$ around $q_{\mathrm{v}}^i$.
    
    The visual attention output for $q_{\mathrm{v}}^i$ is defined as
    \begin{equation}
        \mathrm{Attn}(q_{\mathrm{v}}^i) =
        \mathrm{softmax}\left(
        \frac{q_{\mathrm{v}}^i [\,K_{\mathrm{t}}; \tilde{K}_{\mathrm{v}}^i\,]^\top}{\sqrt{d}}
        \right)
        [\,V_{\mathrm{t}}; \tilde{V}_{\mathrm{v}}^i\,].
    \end{equation}
\end{enumerate}

As shown in Figure~\ref{figure:swin_initial_issue}(b), TVSDA significantly restores the model's ability to generate semantic content; however, the overall layout exhibits severe repetitive patterns.



\subsection{Initial Structural Guidance}
\label{sec:pa_intial_issue}


To preserve a reasonable semantic layout in high-resolution images, an effective approach is to introduce low-resolution structural guidance~\cite{du2024demofusion,kim2025diffusehigh,qiu2024freescale,bu2025hiflow}. This strategy incorporates low-resolution structural priors at the initial stage of the denoising process, and is referred to as Initial Structural Guidance (ISG).


Specifically, given a generated low-resolution image $x_0^L$, we upsample it to obtain $x_{0,up}^H = \texttt{upsample}(x_0^L)$. We then inject noise at an intermediate timestep $\tau \in (0,1)$:

\begin{equation}\label{eq:inject}
x_{\tau}^H = \tau x_1^H + (1 - \tau) x_{0,up}^H,
\end{equation}

where $x_1^H$ denotes Gaussian noise. The value of $\tau$ follows the setting in~\cite{bu2025hiflow} due to its strong empirical performance. The resulting $x_\tau^H$ serves as the starting point of the reverse process, enabling the model to skip early denoising steps while preserving low-frequency structural consistency (see Figure~\ref{figure:swin_initial_issue}(c)).

However, as shown in Figure~\ref{figure:pa_intial_issue}, this strategy suffers from two key limitations:  

\textbf{(1) Local semantic repetition due to limited structural constraint.} 
Since the structural prior is injected only at the initial stage, it is not maintained throughout the reverse process, causing the global semantic layout to gradually drift and often resulting in repetitive structures.  

\textbf{(2) Details blurred due to low-frequency trajectory.} 
Eq.~\ref{eq:inject} defines a transition trajectory starting from a blurred image, where $x_{0,up}^H$ lacks high-frequency components, and progressively evolves into Gaussian noise. As high-frequency information is absent throughout this trajectory, the reverse process tends to reconstructing predominantly low-frequency content, resulting in clean yet over-smoothed outputs with missing fine-grained details.


\subsection{Structure--Detail Guidance}

To overcome these issues, we further impose structural and detail guidance during inference. Specifically, we aim to obtain a new velocity field $\hat{v}^{\text{H}}_\theta(t_i)$ at each timestep $t_i$ such that the inversion process (Eq.~\ref{eq:v-theta_3}) ultimately generates high-resolution images $x^{H}_{0}$ with coherent structures and rich details. To guide the velocity refinement, we consider the single-step predicted clean image at time $t_i$:
\begin{equation}
x^{H}_{0 \gets t_i} = x^{H}_{t_i} + v^{H}_{t_i}(0 - t_i)
\end{equation}
where $x^{H}_{0 \gets t_i}$ denotes the predicted clean image at time $t_i$ with $v^{H}_{t_i}$ fixed. To this end, we encourage each $x^{H}_{0 \gets t_i}$ to be structurally consistent and detail-rich, providing step-wise feedback to refine the velocity field throughout the entire inference trajectory.  Since the velocity field governs the evolution trajectory toward $x_0$, improving these intermediate predictions leads to a more accurate and detail-preserving final result.

\begin{figure}[t]
    \centering
    \includegraphics[width=\linewidth]{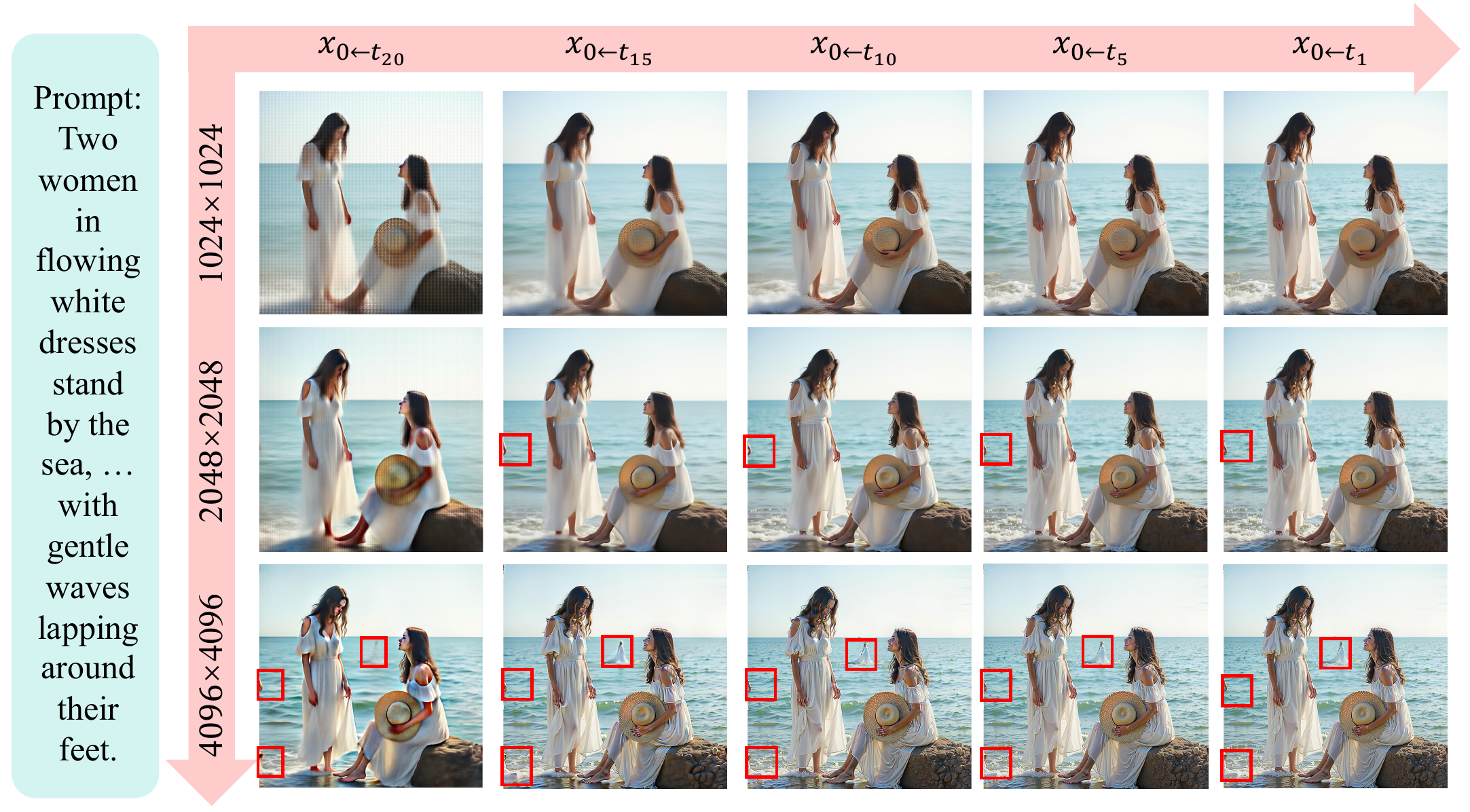}
    \vspace{-0.25cm}
    \caption{Single-step predicted clean image across different scales and time steps. 
    The red boxes highlight regions with repetitive patterns. \textbf{Zoom in} for better visualization.}
    \label{figure:cross_scale_time_motivation}
    \vspace{-5pt}
\end{figure}

Therefore, we investigate the variation of structural and detailed information across different resolutions and time steps $t_i$.
As illustrated in Figure~\ref{figure:cross_scale_time_motivation}, we make three key observations:
\textbf{(1) lower-resolution scales reduce repetitive patterns and produce more reasonable global semantic layouts;}
\textbf{(2) higher-resolution scales generate richer details with sharper textures;}
and \textbf{(3) as the denoising process progresses (i.e., as $t_i$ decreases), fine-grained details gradually become clearer.}

\begin{figure}
    \centering
    \includegraphics[width=0.98\linewidth]{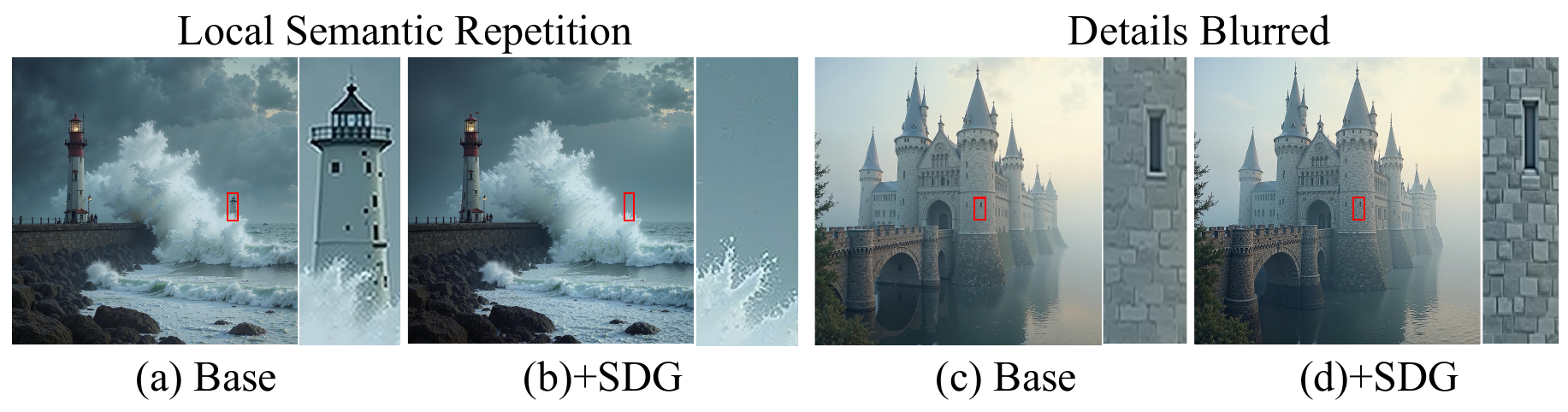}
    \vspace{-0.25cm}
    \caption{Local semantic repetition. 
Base and blurred details denotes the method that uses only text-visual scale-decoupled attention (TVSDA) and initial structural guidance (ISG). \textbf{Zoom in} for better visualization.
 } 
    \label{figure:pa_intial_issue}
    \vspace{-5pt}
\end{figure}

Based on the above observations, we propose a \textbf{structure--detail guidance strategy (SDG)} to promote structural coherence while enhancing fine-grained details from two perspectives: \textbf{cross-scale guidance} and \textbf{cross-temporal guidance}. Figure~\ref{figure:pa_intial_issue} illustrates that, with the corrections from SDG, we are able to maintain coherent structural layouts while simultaneously improving the sharpness of fine-grained details.

\subsubsection{Cross-Scale Guidance}
\label{sec:CSG}

For cross-scale guidance, we formulate the following principle:
\textbf{the low-frequency components (semantic layout) of the generated image should align with those of the low-resolution image to preserve structural coherence,
while the high-frequency components (fine details) should align with those of the high-resolution image to retain rich textures.} Based on this principle, we introduce the following cross-scale guidance (CSG):

\begin{equation}\label{eq:sgaglin}
\begin{aligned}
\Delta v^{\text{H}}_{csg,t_i} 
&= -\frac{1}{t_i}  \Big(\underbrace{\mathcal{F}_{lp}(x^{\text{L}}_{0 \gets t_i}) - \mathcal{F}_{lp}(x^{\text{H}}_{0 \gets t_i})}_{\text{low-frequency correction}} \\
&\quad + \underbrace{\mathcal{F}_{hp}(x^{\text{H}}_{0 \gets t_i}) - \mathcal{F}_{hp}(x^{\text{L}}_{0 \gets t_i})}_{\text{high-frequency correction}}\Big),
\end{aligned}
\end{equation}

where $\mathcal{F}_{lp}$ denotes a low-pass filter with normalized cutoff frequency $W$ (set to 0.4 in our experiments), 
and  $\mathcal{F}_{hp}$ denotes its complementary high-pass filter. The velocity increment $\Delta v^{\text{H}}_{csg,t_i}$ induces a single-step clean image correction
$\Delta x^{\text{H}}_{0\gets t_i} = (0-t_i)\Delta v^{\text{H}}_{csg,t_i}$, 
aligning low-frequency components with the low-resolution image and high-frequency components with the high-resolution image.


\subsubsection{Cross-Temporal Guidance}

In cross-temporal guidance, we observe that the single-step predicted clean image gradually becomes sharper as the denoising process progresses. Therefore, we define a similar correction term to further enhance fine-grained details:
\begin{align}\label{eq:tgaglinlp}
\Delta v^{\text{H}}_{ctg,t_i} 
&= -\frac{1}{t_i}  \Big(\underbrace{\mathcal{F}_{hp}({x}^{\text{H}}_{0 \gets t_{i}}) - \mathcal{F}_{hp}({x}^{\text{H}}_{0 \gets t_{i+1}})}_{\text{high-frequency correction}}\Big). 
\end{align}

By computing the difference in high-frequency components between consecutive time steps, 
the velocity increment $\Delta v^{\text{H}}_{tg,t_i}$ captures the temporal evolution of fine details.

\subsubsection{Velocity Correction}
Leveraging the two guidance terms, the refined $\hat{v}^{\text{H}}_{t_i}$ is defined as:
\begin{equation}\label{eq:recomputed-velocity}
\hat{v}^{\text{H}}_{t_i} = v^{\text{H}}_{t_i}+ \alpha_{t_i} \left(\Delta v^{\text{H}}_{csg,t_i} + \Delta v^{\text{H}}_{ctg,t_i}\right),
\end{equation}
where $v^{\text{H}}_{t_i}$ is the original velocity prediction, and $\alpha_{t_i} = t_i / \tau$ is a time-dependent weight that gradually weakens the guidance strength.



\section{Experiments}
\noindent\textbf{\textit{Experimental Settings.}}
Following prior work~\cite{du2024max,bu2025hiflow}, we conduct experiments with Flux.1.0-dev~\cite{flux2024} at three resolutions: 2048$\times$2048, 2048$\times$4096, and 4096$\times$4096. All experiments are implemented in PyTorch and run on an NVIDIA RTX 8000 GPU.

\begin{figure*}
    \centering
    \includegraphics[width=0.95\linewidth]{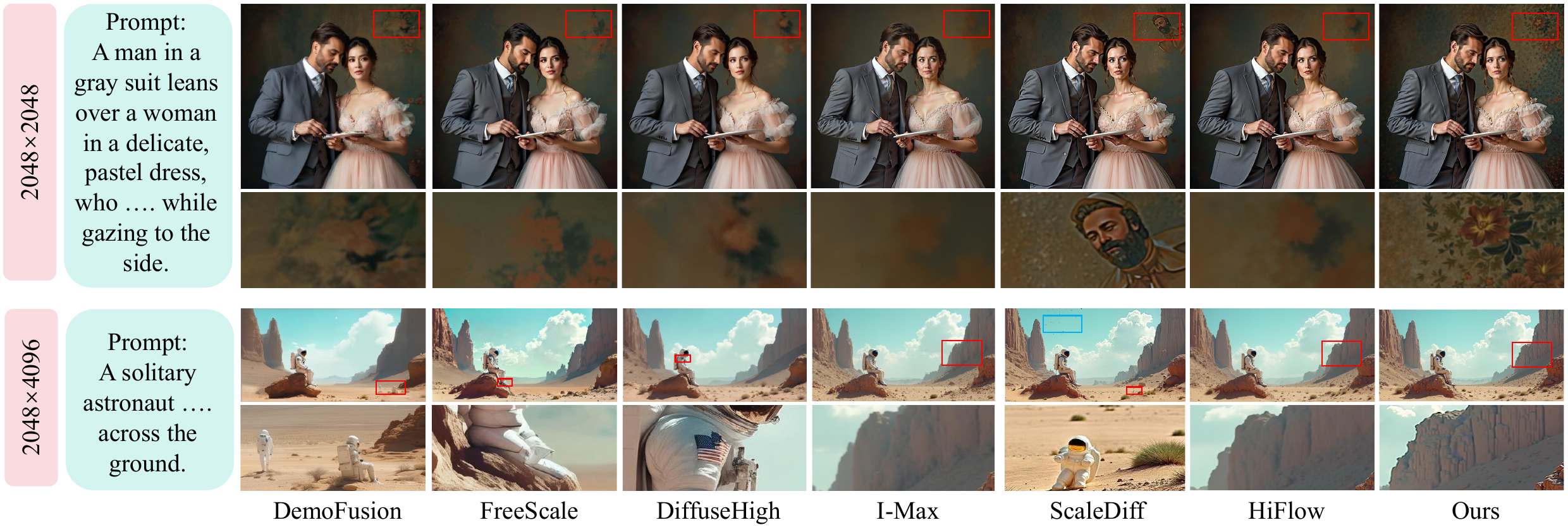}
    \vspace{-0.25cm}
    \caption{Qualitative comparison with other methods on $2048 \times 2048$ and $2048 \times 4096$. Red boxes highlight regions where our method produces clearer and more detailed results, while the blue box indicates dot-like artifacts generated by ScaleDiff. \textbf{Zoom in} for better visualization.}

    \label{figure:2k_2k_vis}
    \vspace{-5pt}
\end{figure*}

\begin{figure*}
    \centering
    \includegraphics[width=0.95\linewidth]{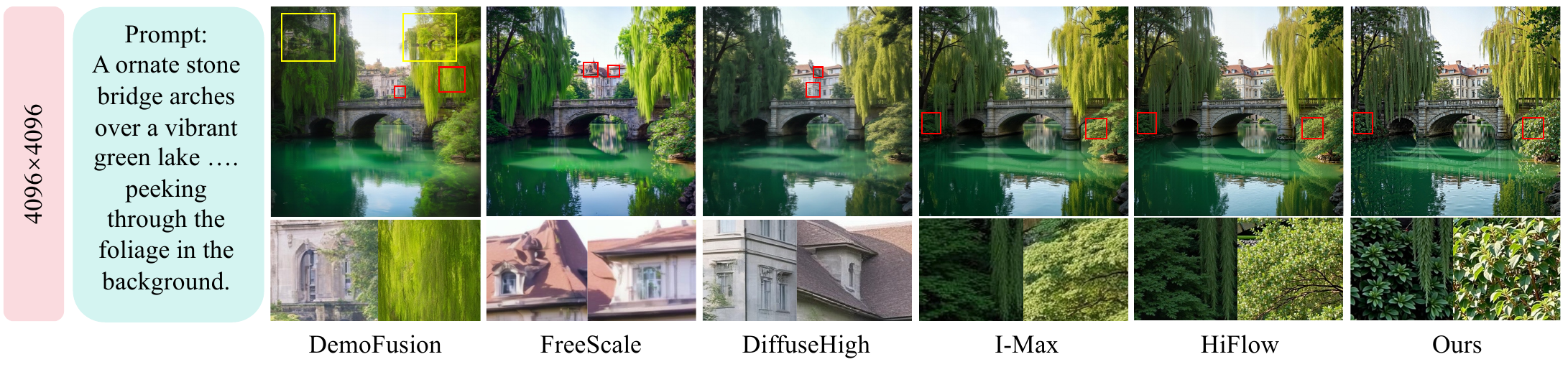}
    \vspace{-0.25cm}
    \caption{Qualitative comparison with other methods on $4096 \times 4096$. Red boxes highlight regions where our method produces clearer and more detailed results while yellow boxes mark the localized repetitive tiny structures produced by DemoFusion. \textbf{Zoom in} for better visualization.}

    \label{figure:4k_4k_vis}
    \vspace{-5pt}
\end{figure*}
\noindent\textbf{\textit{Evaluation.}}
Following previous works~\cite{du2024demofusion,qiu2024freescale,kim2025diffusehigh}, we collect 1K high-quality captions spanning diverse scenarios for image generation. For quantitative evaluation, we adopt widely used metrics including FID~\cite{heusel2017gans}, KID~\cite{binkowski2018demystifying}, and CLIP Score~\cite{radford2021learning}. FID and KID are computed between generated images and a reference set of 10K high-quality real images sourced from LAION-High-Resolution~\cite{schuhmann2022laion}.

However, since FID and KID require resizing images to $299 \times 299$, they may fail to capture fine-grained details in high-resolution generation, and thus often do not fully reflect perceptual quality~\cite{du2024demofusion,jeong2025latent}. To address this limitation, following~\cite{kim2025diffusehigh,qiu2024freescale}, we compute FID and KID at the patch level, denoted as FID$_p$ and KID$_p$, respectively, providing a more reliable assessment of high-resolution image quality.

\subsection{Comparison of Results}
High-resolution image generation methods can be broadly categorized into three groups: training-free diffusion models, super-resolution models, and training-based methods. In the main experiments, we primarily compare our method with training-free diffusion models, while super-resolution and training-based methods are evaluated in the ablation studies. For training-free diffusion models, we consider several state-of-the-art approaches, including three U-Net-based methods: (i) DemoFusion~\cite{du2024demofusion}, (ii) FreeScale~\cite{qiu2024freescale}, and (iii) DiffuseHigh~\cite{kim2025diffusehigh}; and three DiT-based methods: (iv) I-Max~\cite{du2024max}, (v) ScaleDiff~\cite{koh2025scalediff}, and (vi) HiFlow~\cite{bu2025hiflow}.

\begin{table*}[t]
\centering
\resizebox{\textwidth}{!}{
\small
\begin{tabular}{c|ccccc|ccccc|ccccc}
\toprule
\multirow{2}{*}{Method}
 & \multicolumn{5}{c|}{$2048 \times 2048$} & \multicolumn{5}{c|}{$2048 \times 4096$} & \multicolumn{5}{c}{$4096 \times 4096$} \\
\cmidrule(lr){2-6} \cmidrule(lr){7-11} \cmidrule(lr){12-16}

& FID($\downarrow$)& KID($\downarrow$) & FID$_p$ ($\downarrow$) & KID$_p$ ($\downarrow$) & CLIP Score ($\uparrow$)
& FID($\downarrow$)& KID($\downarrow$) & FID$_p$ ($\downarrow$) & KID$_p$ ($\downarrow$) & CLIP Score ($\uparrow$)
& FID($\downarrow$)& KID($\downarrow$) & FID$_p$ ($\downarrow$) & KID$_p$ ($\downarrow$) & CLIP Score ($\uparrow$)\\
\midrule

DemoFusion~\cite{du2024demofusion} (CVPR'24)  & 49.439 & 0.007 & 18.448 & 0.005 & \underline{33.93}
      & 51.184 & 0.008 & 26.251 & 0.007 & 32.84
      & 55.593 & 0.010 & 33.120 & 0.010 & 33.65 \\

FreeScale~\cite{qiu2024freescale} (ICCV'25)   & 44.651 & 0.004 & 13.221 & 0.003 & 33.80
      & 47.342 & 0.004 & 22.418 & 0.004 & 32.90
      & 48.013 & 0.005 & 24.809 & 0.006 & \underline{33.68} \\

DiffuseHigh~\cite{kim2025diffusehigh}  (AAAI'25) & 47.671 & 0.005 & 18.668 & 0.005 & 33.60
      & 50.094 & 0.006 & 28.150 & 0.009 & 32.92
      & 51.738 & 0.007 & 35.169 & 0.012 & 33.59 \\

I-max~\cite{du2024max} (Arxiv'24) & 43.125 & 0.003 & 12.763 & 0.003 & 33.61
      & 45.132 & 0.005 & 19.049 & \underline{0.004} & 32.89
      & 46.872 & 0.005 & 22.815 & 0.007 & 33.57 \\

ScaleDiff~\cite{koh2025scalediff} (NeurIPS'25)  & \underline{41.900} & 0.003 & 12.345 & \textbf{0.002} & 33.70
      & 43.919 & 0.005 & 16.955 & \underline{0.004} &  \textbf{33.02}
      & \multicolumn{5}{c}{OOM} \\

HiFlow~\cite{bu2025hiflow} (NeurIPS'25)    & 42.180 & \underline{0.002} & \underline{11.546} & \textbf{0.002} & 33.55
      & \underline{42.987} & \underline{0.003} & \underline{15.920} & \underline{0.004} & 32.87
      & \underline{43.832} & \underline{0.003} & \underline{18.468} & \underline{0.005} & 33.52\\

Ours  & \textbf{39.175} & \textbf{0.001} & \textbf{10.362} & \textbf{0.002} & \textbf{34.01}
      & \textbf{39.454} & \textbf{0.001} & \textbf{12.304} & \textbf{0.003} & \underline{32.96}
      & \textbf{39.687} & \textbf{0.001} & \textbf{14.047} & \textbf{0.003} & \textbf{33.69} \\

\bottomrule
\end{tabular}
}
\caption{Quantitative comparison under different resolution settings. OOM indicates that the method runs out of GPU memory. We denote the best results in bold and the second-best results with \underline{underline}. $\downarrow$ indicates that lower values are better, while $\uparrow$ indicates that higher values are better.}
\label{tab:quantitative_results}
 \vspace{-0.5cm}
\end{table*}

\noindent\textbf{Visual Comparison.}  We first present qualitative comparisons of different methods in Figure~\ref{figure:2k_2k_vis} and Figure~\ref{figure:4k_4k_vis}, and make the following observations:

\textbf{(1)} U-Net-based methods, such as DemoFusion, FreeScale, and DiffuseHigh, tend to produce structurally inconsistent results when generating ultra-high-resolution images. For example, at a resolution of $4096 \times 4096$, these methods exhibit chaotic local structures in buildings. Moreover, DemoFusion is prone to local repetitive artifacts, such as duplicated small objects (e.g., multiple identical astronauts) at $2048 \times 4096$.

\textbf{(2)} Among the DiT-based methods, ScaleDiff tends to produce repetitive patterns, which are observed at both $2048 \times 2048$ and $2048 \times 4096$. In addition, it occasionally introduces dot-like artifacts, as highlighted by the blue boxed region at $2048 \times 4096$.
In contrast, although I-Max and HiFlow do not exhibit such artifacts, their generated results tend to be overly smooth and lack fine details. Compared with these methods, our approach produces images with more coherent structural layouts and significantly richer details.

\noindent\textbf{Quantitative Comparison.} We further present quantitative comparisons to evaluate the performance of different methods. As shown in Table~\ref{tab:quantitative_results}, we have the following observations:

\textbf{(1)} On key visual quality metrics, including FID, KID, FID$_p$, and KID$_p$, our method consistently achieves the best performance across all evaluation settings, demonstrating its superior capability in generating high-quality high-resolution images.

\textbf{(2)} For the image–text alignment metric (CLIP Score), our method achieves competitive performance, being only marginally lower than ScaleDiff under the $2048 \times 4096$ setting. Importantly, ScaleDiff is unable to perform inference at $4096 \times 4096$ due to GPU memory constraints, whereas our method remains stable and produces valid results at this resolution.

\subsection{Ablation Study}
In this section, we provide a comprehensive experimental evaluation of our method, including ablation studies and additional analyses. Unless otherwise specified, all experiments are conducted at a resolution of $4096 \times 4096$.

\begin{figure}
    \centering
    \includegraphics[width=0.95\linewidth]{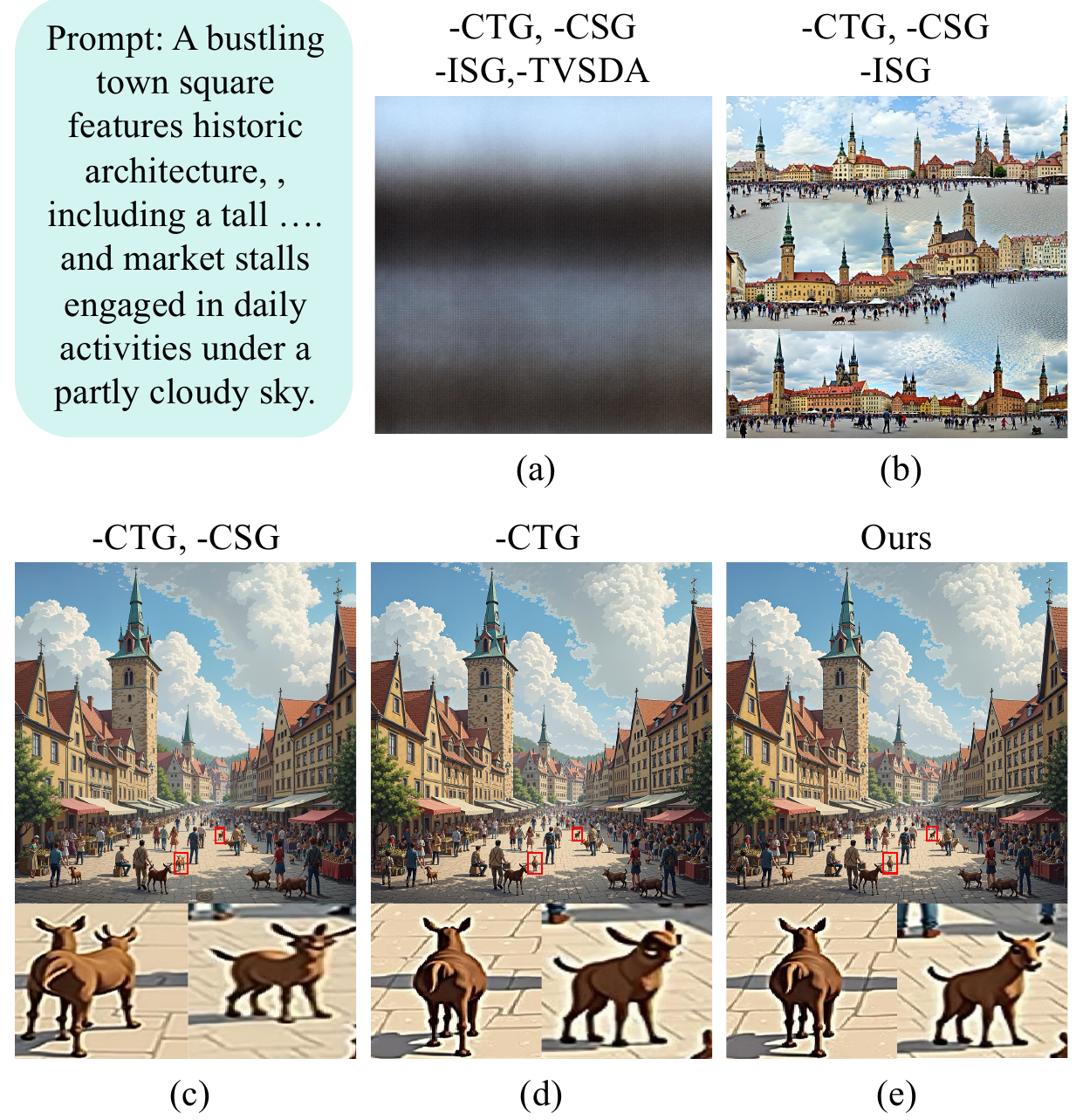}
    \vspace{-0.25cm}
    \caption{Component ablation. We illustrate the effect of sequentially removing different
components of our method on FLUX. 
\textbf{Zoom in} for better visualization.} 
    \label{figure:differnt_component_vis}
    \vspace{-5pt}
\end{figure}

\subsubsection{Effect of Components}

\begin{table}[t]
\centering
\small
\resizebox{\columnwidth}{!}{
\begin{tabular}{
l
S[table-format=3.3]
S[table-format=1.3]
S[table-format=2.3]
S[table-format=1.3]
S[table-format=2.2]
}
\toprule
Method 
& {FID ($\downarrow$)} 
& {KID ($\downarrow$)} 
& {FID$_p$ ($\downarrow$)} 
& {KID$_p$ ($\downarrow$)} 
& {CLIP Score ($\uparrow$)} \\
\midrule
-CTG, -CSG, -ISG, -TVSDA & 228.652 & 0.372 & 202.781 & 0.240 & 12.11 \\
-CTG, -CSG, -ISG         & 107.029 & 0.021 & 33.630 & 0.008 & 29.52 \\
-CTG, -CSG               & 48.709 & 0.006 & 24.009 & 0.009 & 33.53 \\
-CTG                     & 41.744 & 0.003 & 16.976 & 0.004 & 33.65 \\
Ours                     & \bfseries 39.687 & \bfseries 0.001 & \bfseries 14.047 & \bfseries 0.003 & \bfseries 33.69 \\
\bottomrule
\end{tabular}
}
\caption{Ablation study on different components. $\downarrow$ indicates that lower values are better, while $\uparrow$ indicates that higher values are better.}
\label{tab:ablation}
 \vspace{-1cm}
\end{table}

Figure~\ref{figure:differnt_component_vis} presents an ablation study by sequentially removing different components. Comparing Figure~\ref{figure:differnt_component_vis}(a) and (b), we observe that introducing TVSDA significantly improves semantic quality and restores meaningful content, but also leads to layout repetition artifacts.
As shown in Figure~\ref{figure:differnt_component_vis}(c), incorporating ISG improves global layout consistency, while local structural issues and blurred details still remain, e.g., two dogs sharing a single body and unclear facial structures.
Finally, as shown in Figure~\ref{figure:differnt_component_vis}(d) and (e), adding intermediate-state guidance terms (CSG and CTG) progressively alleviates these artifacts, producing more coherent structures and richer details.
Table~\ref{tab:ablation} further shows that progressively adding components consistently improves both global and local metrics.

\subsubsection{Comparison of different attention mechanisms}
\begin{figure}
    \centering
    \includegraphics[width=0.95\linewidth]{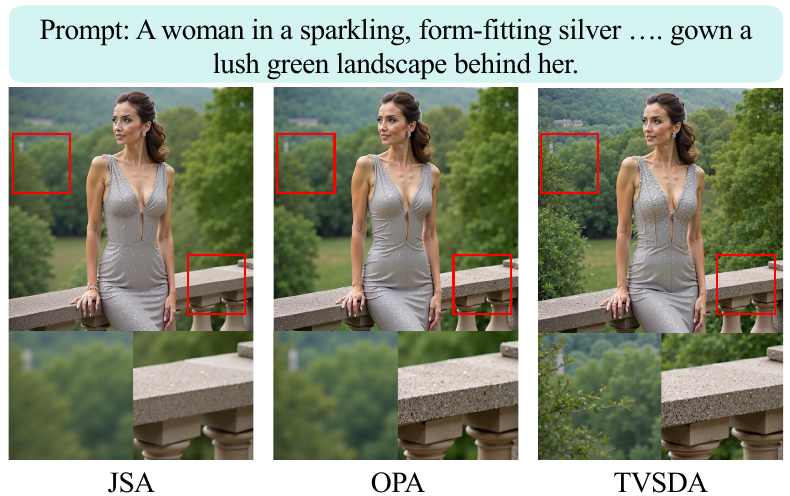}
    \vspace{-0.25cm}
    \caption{Comparison of different attention mechanisms.  
\textbf{Zoom in} for better visualization.} 
    \label{figure:differnt_attention_vis}
    \vspace{-5pt}
\end{figure}
In this section, we evaluate TVSDA by comparing it with two alternative attention schemes: (1) the original joint self-attention (JSA) and (2) an overlapping patch-wise attention strategy (OPA)~\cite{bar2023multidiffusion}. Here, we only adopt the initial structural guidance (ISG)  in all settings to maintain global structural coherence.
Figure~\ref{figure:differnt_attention_vis} compares different attention mechanisms. Both OPA and TVSDA improve fine-grained details over JSA. For example, in the bottom-right red box, JSA produces a blurry fence, whereas OPA and TVSDA generate sharper results.
However, OPA operates on overlapping patches and requires smoothing fusion, introducing blurring artifacts. In contrast, TVSDA avoids patch overlap and yields clearer, more consistent details. For instance, background trees remain blurry with OPA, while our method produces sharper structures.

\subsubsection{Inference Speed}

\begin{table}[t]
\centering
\begin{tabular}{lcccc}
\toprule
Resolution & I-Max & ScaleDiff & HiFlow & Ours \\
\midrule
2048 $\times$ 2048 & 6.7 & 3.8 & 3.9 & \textbf{3.6} \\
2048 $\times$ 4096 & 20.4 & 7.7 & 7.5 & \textbf{6.8} \\
4096 $\times$ 4096 & 71.5 & OOM & 23.7 & \textbf{14.7} \\
\bottomrule
\end{tabular}
\caption{Inference time comparison (minutes  per image) under different generation resolutions. OOM indicates that the method runs out of GPU memory.}
\label{tab:speed_comparison}
\vspace{-0.25cm}
\end{table}
In this section, we compare the inference speed of different DiT-based methods. Table~\ref{tab:speed_comparison} reports the inference time at various generation resolutions. As the resolution increases, all methods incur higher computational costs. Nevertheless, our method consistently achieves the best efficiency across all settings. This speedup mainly comes from replacing global attention with local attention, which substantially reduces the attention computation, especially at high resolutions. Moreover, unlike ScaleDiff, our method remains stable at the highest resolution without out-of-memory issues, demonstrating better scalability for high-resolution image synthesis.

\subsubsection{Comparison of Super-resolution Methods}

\begin{figure}
    \centering
    \includegraphics[width=0.95\linewidth]{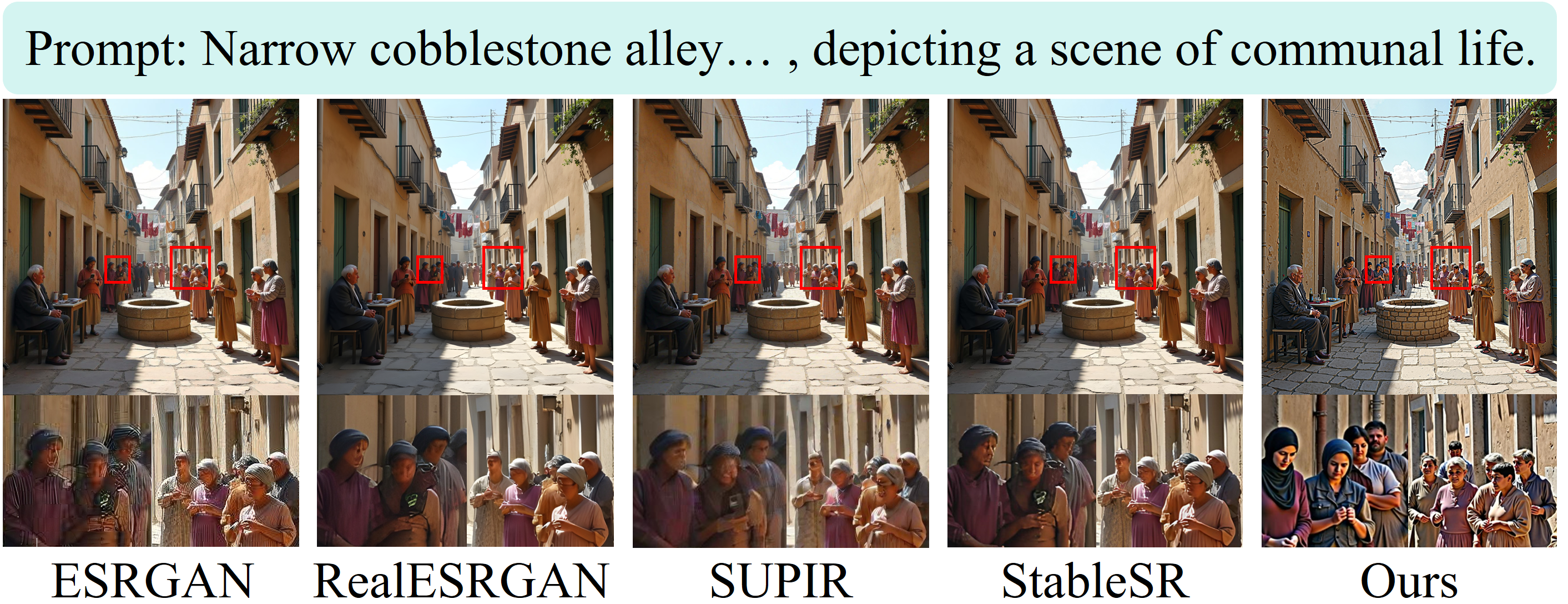}
    \vspace{-0.25cm}
    \caption{Qualitative comparison of different Super-resolution methods. \textbf{Zoom in} for better visualization.}

    \label{figure:super_com_vis}
    \vspace{-5pt}
\end{figure}

In this section, we further compare our method with direct super-resolution approaches, including the GAN-based methods ESRGAN~\cite{wang2018esrgan} and RealESRGAN~\cite{wang2021real}, as well as the diffusion-based methods StableSR~\cite{wang2024exploiting} and SUPIR~\cite{yu2024scaling}.
As shown in Figure~\ref{figure:super_com_vis}, super-resolution methods often exhibit local artifacts, whereas our method produces more natural and coherent results.

\subsubsection{Comparison of Training-based Methods}
\begin{figure}
    \centering
    \includegraphics[width=0.95\linewidth]{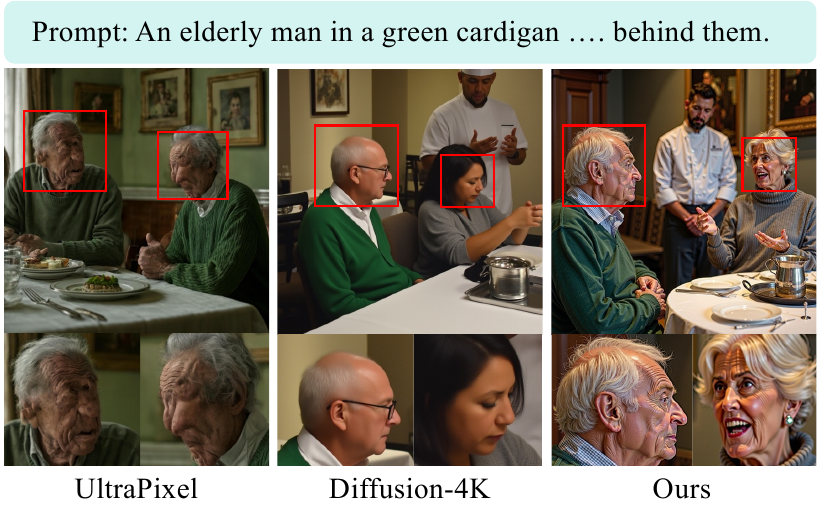}
    \vspace{-0.25cm}
    \caption{Qualitative comparison of different training-based methods. \textbf{Zoom in} for better visualization.}
    \label{figure:training_based_com_vis}
    \vspace{-5pt}
\end{figure}

We compare our method with two training-based approaches, UltraPixel~\cite{ren2024ultrapixel} and Diffusion-4K~\cite{zhang2025diffusion}. As shown in Figure~\ref{figure:training_based_com_vis}, UltraPixel often produces unrealistic content, while Diffusion-4K suffers from blurry details. In contrast, our method generates more natural results.

\subsubsection{Generalization to Other Model Variants}

\begin{figure}
    \centering
    \includegraphics[width=0.95\linewidth]{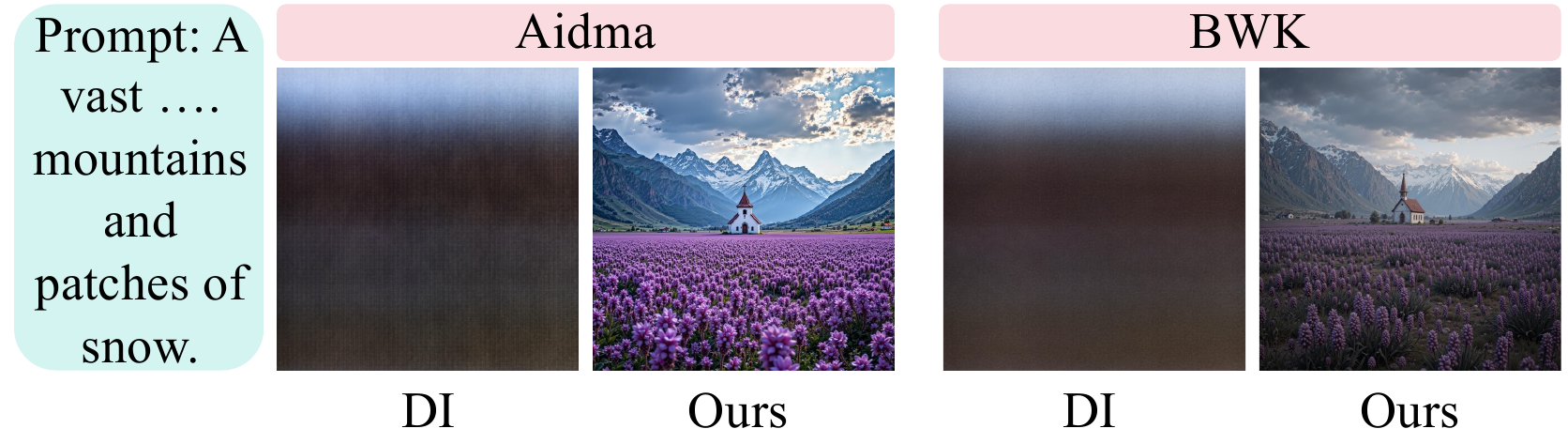}
    \vspace{-0.25cm}
    \caption{Qualitative comparison of other model variants. \textbf{Zoom in} for better visualization.}
    \label{figure:differen_model_vis}
    \vspace{-5pt}
\end{figure}

We further evaluate the generalization ability of our method on several customized T2I models. Specifically, we conduct experiments on Aidma~\cite{flux_pro_detailer_lora} and BWK~\cite{black_myth_wukong_flux_lora}. As shown in Figure~\ref{figure:differen_model_vis}, these variants  suffer from severe semantic degradation under high-resolution inference. Our method can be directly applied to these models and significantly improves the quality of high-resolution image.

\section{Conclusion}
In this paper, we propose a training-free framework for high-resolution image generation with DiT models. Our method introduces text–visual scale-decoupled attention (TVSDA) to reduce semantic degradation by preserving global text perception while adapting image-token receptive fields to the training resolution. We further design structure–detail guidance (SDG), which improves structural coherence and fine details through cross-scale guidance (CSG) and cross-temporal guidance (CTG). Extensive experiments show that our method achieves better structure and detail fidelity than existing approaches.
\section{Acknowledgements}
 This work is supported by the National Natural Science Foundation of China (Grant No. 62276031,62406098).

\bibliographystyle{ACM-Reference-Format}
\balance
\bibliography{sample-base}


\end{document}